# ContourRend: A Segmentation Method for Improving Contours by Rendering [*]

Junwen Chen[1,2], Yi Lu[1], Yaran Chen[1], Dongbin Zhao[1], and Zhonghua Pang[2]

[1]State Key Laboratory of Management and Control for Complex Systems
Institute of Automation, Chinese Academy of Sciences, Beijing 100190, China
University of Chinese Academy of Sciences, Beijing 101408, China
[2]Key Laboratory of Fieldbus Technology and Automation of Beijing,
North China University of Technology, Beijing, 100144, China

**Abstract.** A good object segmentation should contain clear contours and complete regions. However, mask-based segmentation can not handle contour features well on a coarse prediction grid, thus causing problems of blurry edges. While contour-based segmentation provides contours directly, but misses contours' details. In order to obtain fine contours, we propose a segmentation method named ContourRend which adopts a contour renderer to refine segmentation contours. And we implement our method on a segmentation model based on graph convolutional network (GCN). For the single object segmentation task on cityscapes dataset, the GCN-based segmentation contour is used to generate a contour of a single object, then our contour renderer focuses on the pixels around the contour and predicts the category at high resolution. By rendering the contour result, our method reaches 72.41% mean intersection over union (IoU) and surpasses baseline Polygon-GCN by 1.22%.

**Keywords:** Image segmentation, Convolution neural networks, Contour renderer, Graph convolutional network

## 1 Introduction

Convolutional neural network (CNN) methods bring various breakthroughs to the field of computer vision, improve the accuracy in the tasks of image classification [1][2], image classification and location [3], object detection [4], image segmentation [5], and even surpass the human performance. More and more image processing tasks begin to rely on the rich features provided by CNN.

In image segmentation task, semantic segmentation predicts the label of every pixel. And CNN can also easily provide the encoding of segmentation information for kinds of usages. Full convolution network (FCN) [6] uses a fully convolutional struc-ture for segmentation and builds a skip architecture to connect semantic information in different depth of the convolution layers. In FCN, the features 8×, 16×, 32× smaller than the input are used by the transposed convolution to predict mask result. U-Net [7] is also a fully convolutional network and has a symmetric architecture in encoding and decoding feature maps. U-Net concatenates the feature maps with the same resolution in the encoder and decoder and uses transposed convolution to restore these features to output mask results at a higher resolution. In instance segmentation task, segmentation focuses on distinguishing between pixel regions of different objects. Mask R-CNN [8] as the baseline of this task, adds a CNN segmentation branch on Faster R-CNN [9]. Faster R-CNN provides the feature map of the object in a 14×14 grid for the CNN branch, and the CNN branch predicts a 28×28 mask result.

[*] This work is supported partly by National Key Research and Development Plan under Grant No.2017YFC1700106, and National Natural Science Foundation of China under Grant 61673023, Beijing University High-Level Talent Cross-Training Project (Practical Training Plan)

Although these methods utilize the excellent feature extraction ability of the convolution operator, the feature maps with 8 times or 16 times smaller than the input are too coarse for segmentation. While upsampling or resizing these coarse masks to the results with the same size of the input images, there are blurry edges on the mask results' contours which limits the segmentation performance. To reduce this limitation, some methods focus on modifying the convolutional operator and pooling operator to lessen the down-sampling effect in the mask-based models. PSPNet [10] uses pyramid pooling module to fuse the global context information and reduces false positive results. DeepLab family [11][12][13] and DenseASPP [14] use dilated convolution to expand the size of receptive field and improves the resolution of segmentation.

To avoid down-sampling effect in the mask-based models, some contour-based segmentation models that distinguish the object by the contour formed by the contour vertices are proposed. These models can obtain clear contours at the same resolution as the input image by determining the coordinates of the contour vertices. PolarMask [15] learns to predict dense distance regression of contour vertices from the object's center position in a polar coordinate. Polygon-RNN [16] and Polygon-RNN++ [17] utilize RNN to find contour vertices one by one. Curve-GCN [18] implements graph convolutional network (GCN) to obtain the coordinates of the contour vertices by regression. Curve-GCN can simultaneously adjust the coordinates of a fixed number of vertices from the initial contour to the target.

Although the above contour-based segmentation methods avoid the effect by down-sampling and directly restore the resolution, they are unable to provide complex edges due to the limitation of the fixed number of contour vertices. To this end, we propose a segmentation method to reconsider the segmentation process. We focus on improve the contour-based segmentation models by adding a contour renderer, so we call our method ContourRend. Rendering on the segmentation results has been learned in the work PointRend [19], which using mask scores to select the unclear points around the contour. Different from PointRend, the contour renderer of our method directly obtains rendering points by offsetting the contour vertices from the contour-based model, and directly renders on the mask with the same resolution as the input image.

Our contributions in this paper are two folds,

1. We propose a segmentation method to improve the accuracy of contour-based segmentation models by adding a contour renderer, named ContourRend.
2. The experimental results of ContourRend on the single object segmentation task with cityscapes dataset show the improvement both in training and testing. ContourRend reaches 72.41% mean IoU, and surpasses baseline Polygon-GCN by 1.22%.

## 2   Method

Our method ContourRend consists of a contour generator and a contour renderer, completes the segmentation problem in two steps as shown in Fig.1. First, the contour generator generates an initial contour prediction; then, the contour renderer optimizes the contour prediction in pixel level. The contour generator is a contour-based segmentation models which provides the backbone feature map and the initial contour vertex for the contour renderer. The contour renderer optimizes the initial contour like rendering and outputs the mask results with refined edges. Section 3.1 introduces the architecture and the function of the generator, and section 3.2 details how the contour render module refines the initial contour.

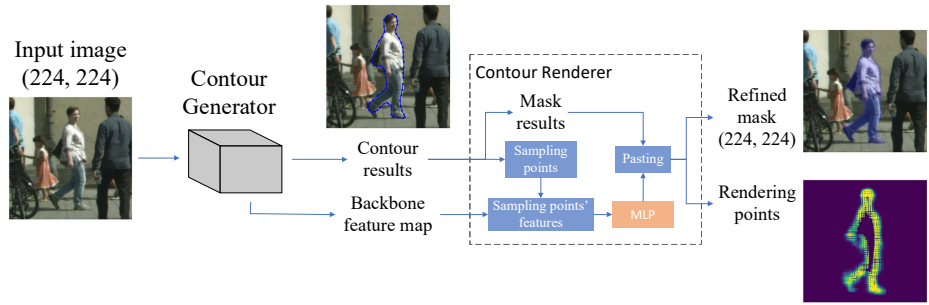

**Fig. 1.** Inference process of ContourRend. The contour generator provides contour results and the backbone feature map for the contour renderer, and the contour renderer optimizes the contour results by using a MLP to classify the sampled points around the contour.

### 2.1   Contour generator

Contour generator aims to generate the backbone feature map and contour vertices for contour renderer. And we build our contour generator according to Tian's GCN-based segmentation model [20] which is similar to Curve-GCN. Graph neural network (GNN) is powerful at dealing with graph structure data and exploring the potential relationship, and using GCN in the mask-based model could improve the features' expression [21] and the result's accuracy [22]. Tian and Curve-GCN utilize GCN to predict contour vertices, and Tian's model implements DeepLab-ResNet to provide the backbone feature map. Tian uses the model on magnetic resonance images and outperforms several state-of-the-art segmentation methods. Fig. 2. shows the architec-ture of our contour generator. The DeepLab-ResNet provides a 512×28×28 backbone feature map, and two branches of the networks consist of a 3×3 convolution layer and a fully connected layer after the backbone, respectively. The two branches provide a 1×28×28 edge feature map and a 1×28×28 vertex feature map, and concatenate the backbone feature map. Before the GCN modules, a 3×3 convolution layers processes the 514×28×28 feature map provided by the backbone and the two branches, and outputs the 320×28×28 feature map.

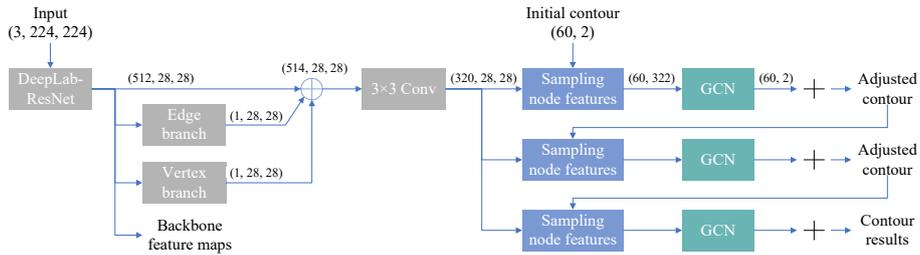

**Fig. 2.** Contour generator's architecture.

The GCN module use a fixed topology graph to represent the contour as same as the Curve-GCN. The relationship between nodes and edges can be regarded as a ring composed of nodes. Each node is connected to two adjacent nodes on the left side and two on the right side. Fig. 3. illustrates the graph by an example with eight nodes.

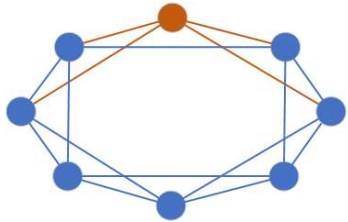

**Fig. 3.** Example of GCN's fixed topology graph. Every node connects with other four adjacent points.

Node features are composed of corresponding coordinate positions on the contour and the contour vertex features extracted from the 320×28×28 feature map, the contour vertex features are extracted by bilinear interpolation according to their 0~1 positions. After GCN propagates and aggregates nodes' features on the graph, the output of the nodes are offsets of the contour vertices. By scaling result points' coordinates from 0~1 to the input size, the contour generator simply obtains contour segmentation results with the same resolution as the input. And the contour generator is trained by point matching loss (MLoss) according to Curve-GCN. During our training, we calculate the L2 loss, and the predicted contour vertices and the target vertices are both sampled to $K$ points in a clockwise order. And the $p$ and $p'$ are the sets of $K$ predicted points and *K* target points represented by the *x* and *y* coordinates from 0 to 1. The loss function is shown as:

$$L_{\text{match}}(p, p') = \min_{j \in [0,\cdots,K-1]} \sum_{i=0}^{K-1} \left\| p_i - p'_{(j+i)\%K} \right\|_2 \quad (1)$$

where $p_i$ represents the *i* th predicted points, $p'_{(j+i)\%K}$ represents the matching point of $p_i$ while the index offset is *j*, the % indicates modulus operation, and $\left\| p_i - p'_{(j+i)\%K} \right\|_2$ indicates the L2 distance between $p_i$ and $p'_{(j+i)\%K}$.

Finally, the contour generator provides 60 contour vertices and the 512×28×28 backbone feature map for the contour renderer.

## 2.2   Contour renderer

The contour renderer samples points based on the contour vertices provided by the contour generator, extracts the points' features by bilinear interpolation according to their positions, then predicts the category scores of these sampled points by a multi-layer perceptron (MLP) consisted of a 1×1 convolution layer, and finally gets the refine mask result by pasting the sample point categories to the initial contour.

For the process of sampling points, we develop two methods to select points for the contour renderer during training and testing respectively. During the training, contour vertices are used to represent segmentation results, as opposed to the case of mask scores, we can naturally get random points around edges by offsetting the contour results. The output of the contour generator is a fixed number of points in the range 0~1. We randomly offset the *x* and *y* coordinates by -0.09~0.09 to generate *n* offset points for each output points. Then we sample the points' targets from the mask represented by contour results and use cross entropy loss as the loss of the renderer. The contour generator and the contour renderer both use points' features, and the renderer loss can be viewed as an auxiliary loss. Fig. 4. shows the process of the contour renderer during the training. During the testing, there is no need to calculate the gradient, so more points are used to obtain a dense prediction. For every contour vertex, an $N \times N$ ($N \geq 1$) grid is generated, and the contour vertex is located at the center of the grid. $N^2$ points evenly cover a $s \times s$ ($s \in [0, 1]$) square area with the gap of $s/(N-1)$ in both *x* and *y* coordinates.

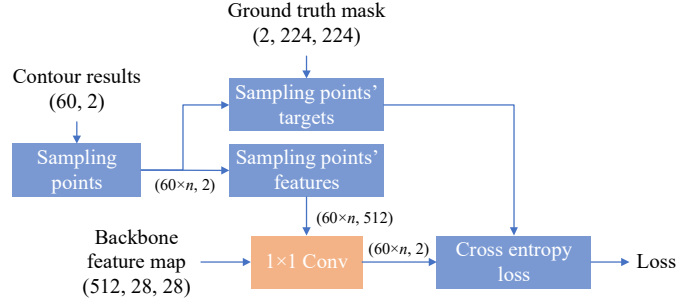

**Fig. 4.** Contour renderer in training. The ground truth mask has two categories (background and foreground).

Then, the renderer optimizes segmentation results by reclassifying single points around the output contour vertices of the generator. Specifically, we input (60×N×N, 512) point features to the MLP, and MLP predicts the (60×N×N, 2) category scores (background and foreground scores) of the corresponding points. We change the contour result of the contour generator into a mask result with the same size as input, and paste the contour renderer's generated points to the mask. Thus, we restore the high resolution while retain the complex edge's details. Fig. 5. shows the process of the contour renderer during the testing.

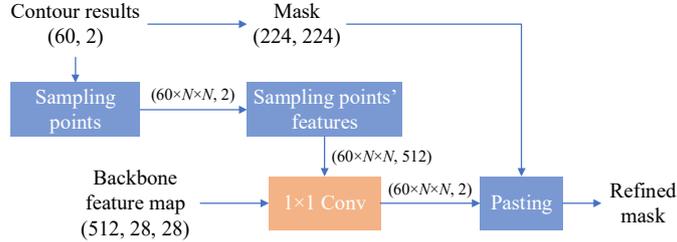

**Fig. 5.** Contour renderer in testing.

## 3  Experiments

We conduct a contrast experiment and an ablation experiment to verify the advantages of our method and test the contour renderer's effect in single object segmentation task on cityscapes dataset. In the contrast experiment, we train our ContourRend and compare with other contour-based segmentation methods. Furthermore, we train our contour generator separately as an ablation experiment to explore the contour renderer's effect.

### 3.1 Dataset

For single object segmentation task, we use the cityscapes dataset and have the same data preprocessing as Curve-GCN [18]. The input is a 224×224 single object image with background, and the object is in the center of the image. The dataset is divided into train set 45984 images, validation set 3936 images and test set 9784 images. And our contour generator's goal is to match the target contour vertices of the single object, our contour renderer's goal is to correctly classify the sampled points' categories (background and foreground).

### 3.2 Implementation

For the contour renderer, we randomly sample 3 ($n = 3$) rendering points around every vertex of the contour generator's result in training process, and use a 1×1 convolution layer to classify the 512 dimension features to 2 categories (background and foreground). In testing process, we select 15×15 ($N = 15$) grid rendering points with the size of 0.09×0.09 ($s = 0.09$ ), and if the foreground's scores of the renderer's result points are higher than 0.3, the points are considered as the foreground points.

We train the entire network end-to-end on four GTX 1080 Ti GPUs with batch size of 8, set the learning rate begin with 3e-4, and 0.1 learning rate decay every 10 epochs. And we use 1e-5 weight decay [23] instead of dropout to prevent overfitting.

### 3.3 Results

#### 3.3.1 Contrast experiment

In the contrast experiment, we train our ContourRend and compare the IoU by categories and the mean IoU with other contour-based methods, Polygon-RNN++ and Polygon-GCN. Table 1 shows the results. Lines 1, 2 refer to Polygon-RNN++ [17], Line 3 refers to Curve-GCN's Polygon-GCN [18]. Our contour generator is similar to Curve-GCN's Polygon-GCN, so we choose Polygon-GCN as our experiment's baseline. From the results, our method surpasses the Polygon-GCN by 1.22%.

**Table 1.** Results of the contrast experiment.

| Methods | Bicycle | Bus | Person | Train | Truck | Motorcycle | Car | Rider | Mean |
|---|---|---|---|---|---|---|---|---|---|
| Polygon-RNN++[17] | 57.38 | 75.99 | 68.45 | 59.65 | 76.31 | 58.26 | 75.68 | 65.65 | 67.17 |
| Polygon-RNN++(with BS)[17] | 63.06 | 81.38 | 72.41 | 64.28 | 78.90 | 62.01 | 79.08 | 69.95 | 71.38 |
| Polygon-GCN 18 | 63.68 | **81.42** | 72.25 | 61.45 | **79.88** | 60.86 | 79.84 | 70.17 | 71.19 |
| CountourRend (ours) | **65.18** | 80.90 | **74.16** | **64.40** | 78.26 | **63.30** | **80.69** | **72.36** | **72.41** |

Fig. 6. shows some of the results to visualize the renderer's effect. The first column is the 224×224 input image, the second column is the contour result of the contour generator which is trained separately, the third column is the ground truth contour, the fourth column is the contour result of the contour generator in ContourRend which is trained with the contour renderer, the fifth column is the mask result predicted by the contour renderer in ContourRend, the sixth column is the ground truth mask and which we use to calculate IoU, the last column is a visualization of the contour renderer's output. For the man in the first line's pictures, the separately trained contour generator can not fit the man's feet and back well (column 2), and the contour generator trained in ContourRend predicts a better contour result (column 4), then after the contour renderer, the man's shoes can also be segmented (column 5). Besides, ContourRend can also predict the women's bag in line 2 column 4, 5 and the truck's tyre in line 3 column 4, 5. ContourRend improves the original contour generator's performance and also refines the details by the contour renderer.

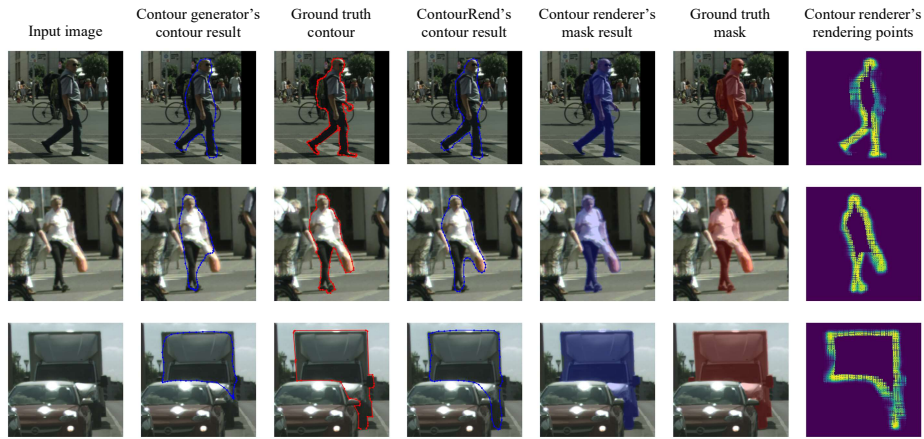

**Fig. 6.** Contrast experiment results. Column 1: Input image, column 2: Contour generator's contour result (trained separately), column 3: Ground truth contour, column 4: ContourRend's contour result, column 5: Rendered mask, column 6: Ground truth mask, column 7: Contour Renderer's point results. ContourRend improves the segmentation result by refining the details around the contour.

### 3.3.2 Ablation experiment

In the ablation experiment, we train the contour generator separately and calculate the IoU by converting the contour result to mask result, then split the contour generator in ContourRend which has been trained in the contrast experiment and also calculate the IoU by it's contour result. Table 2. line 1 shows the result of the contour generator which is not trained with our contour renderer, and line 2 is the result of ContourRend's contour generator which is trained with our contour renderer in the contrast experiment, line 3 is ContourRend's result after the contour renderer improves the contour. From the results of line 1 and line 2, the contour renderer improves the contour generator's mean IoU by 2.67% in the training, and compare with the line 1 and line 3, ContourRend makes 7.16% improvement on the mean IoU by improving the contour in the testing. According to the ablation experiment, our method improves both the contour-based model's accuracy in the training and testing.

Table 2. Results of the ablation experiment.

| Methods | Bicycle | Bus | Person | Train | Truck | Motorcycle | Car | Rider | Mean |
|---|---|---|---|---|---|---|---|---|---|
| Contour Generator | 57.74 | 73.71 | 66.76 | 56.52 | 71.70 | 56.14 | 75.42 | 64.02 | 65.25 |
| ContourRend-ablation | 59.69 | 76.67 | 69.93 | 59.77 | 75.14 | 57.03 | 77.19 | 67.91 | 67.92 |
| ContourRend | **65.18** | **80.90** | **74.16** | **64.40** | **78.26** | **63.30** | **80.69** | **72.36** | **72.41** |

The contour renderer can improve the accuracy during training, because the renderer resamples the segmented pixels and makes the model focus on the object's contour. This phenomenon has also led to a conjecture that the evenly participation of all pixels in the original image in the training may cause computational waste and even lead to the decline of segmentation accuracy. The renderer loss reset the weights of the pixels to participate the segmentation which improves the performance of the baseline model. This can deduce that the pixels around contour play more important role in segmentation than other pixels.

## 4  Conclusion

In order to tackle the problem of blurry edges in mask-based segmentation models' results and improve the accuracy of contour-based segmentation models, we propose a segmentation method by combining a contour-based segmentation model and a contour renderer. In the single object segmentation task on cityscapes dataset, our method reaches 72.41% mean IoU and surpasses Polygon-GCN by 1.22%. And the proposed contour renderer enhanced contour-based segmentation mechanism is also effective to improve the performance of other kinds of contour based segmentation methods, as PolarMask and Curve-GCN.